%% file: main.tex
\pdfoutput=1
\documentclass[11pt]{article}
\input{macros.tex}
\input{shorthand.tex}

\title{
    Logical forms complement probability \\
    in understanding language model (and human) performance
}

\author{Yixuan Wang \\
  University of Chicago \\
  \texttt{yixuanwang@uchicago.edu} \\\And
  Freda Shi \\
  University of Waterloo \\
  Vector Institute, Canada CIFAR AI Chair \\
  \texttt{fhs@uwaterloo.ca} \\}

\begin{document}

\setlength{\Exlabelsep}{0em}
\setlength{\SubExleftmargin}{1em}

\maketitle
\begin{abstract}
  With the increasing interest in using large language models (LLMs) for planning in natural language, understanding their behaviors becomes an important research question.
  This work conducts a systematic investigation of LLMs' ability to perform logical reasoning in \textit{natural language}.
  We introduce a controlled dataset of hypothetical and disjunctive syllogisms in propositional and modal logic and use it as the testbed for understanding LLM performance.
  Our results lead to novel insights in predicting LLM behaviors: in addition to the probability of input \citep{gonen-etal-2023-demystifying,mccoyEmbersAutoregressionShow2024}, logical forms should be considered as important factors.
  In addition, we show similarities and discrepancies between the logical reasoning performances of humans and LLMs by collecting and comparing behavioral data from both.
\end{abstract}

\input{chapters/01-intro}
\input{chapters/02-related}
\input{chapters/03-dataset}
\input{chapters/04-experiments}
\input{chapters/05-human}
\input{chapters/06-discussion}
\input{chapters/07-closing}

\bibliography{custom}

\appendix

\input{chapters/a1-responsible}
\input{chapters/a2-sample}
\input{chapters/a3-extra}

\end{document}

%% file: macros.tex
\usepackage[preprint]{acl}

\usepackage{times}
\usepackage{latexsym}
\usepackage[T1]{fontenc}
\usepackage[utf8]{inputenc}
\usepackage{microtype}
\usepackage{inconsolata}

\usepackage{xcolor}

\usepackage{import}

\usepackage{amsmath}
\usepackage{amssymb}
\usepackage{bbm}
\usepackage{mathtools}
\usepackage[normalem]{ulem}

\usepackage{graphicx}
\graphicspath{{./asset/figure}}
\usepackage{subcaption}

\usepackage{linguex}
\usepackage{enumitem}
\usepackage{booktabs}
\usepackage{siunitx}
\usepackage{multirow}
\usepackage{array}
\usepackage{float}
\usepackage{caption}

\newcolumntype{L}{>{$}l<{$}} 
  \newcolumntype{C}{>{$}c<{$}} 
\newcolumntype{R}{>{$}r<{$}} 

\usepackage[capitalize,noabbrev]{cleveref}
\crefname{ExNo}{Example}{Examples}
\crefname{SubExNo}{Example}{Examples}

\creflabelformat{SubExNo}{(#2#1#3)}
\creflabelformat{ExNo}{(#2#1#3)}
\crefrangelabelformat{SubExNo}{(#3#1#4--#5\crefstripprefix{#1}{#2}#6)}
\crefrangelabelformat{ExNo}{(#3#1#4)--(#5#2#6)}

\crefname{chapter}{Chapter}{Chapters}
\crefname{section}{\S}{\S\S}
\Crefname{section}{\S}{\S\S}
\crefname{table}{Table}{Tables}
\crefname{figure}{Figure}{Figures}
\crefname{algorithm}{Algorithm}{}
\crefname{equation}{Eq.}{}
\crefname{appendix}{Appendix}{}
\crefformat{section}{\S#2#1#3}
\crefname{lemma}{Lemma}{Lemmas}
\crefname{proposition}{Proposition}{Propositions}
\crefname{definition}{Definition}{Definitions}
\crefname{corollary}{Corollary}{Corollaries}
\crefname{example}{Example}{Examples}
\crefname{problem}{Problem}{Problems}
\crefname{remark}{Remark}{Remarks}

\newcommand{\interalia}[1]{\citep[\textit{inter alia}]{#1}}

\usepackage{todonotes}

\definecolor{tticblue}{RGB}{0, 94, 184}

\makeatletter
\ifacl@anonymize
  \newcommand{\blind}[1]{}
\else
  \newcommand{\blind}[1]{#1}
\fi
\makeatother

%% file: shorthand.tex
\newcommand{\modal}{\mathfrak{M}}

%% file: chapters/01-intro.tex
\section{Introduction}

\begin{figure}[!t]
    \includegraphics[width=0.47\textwidth]{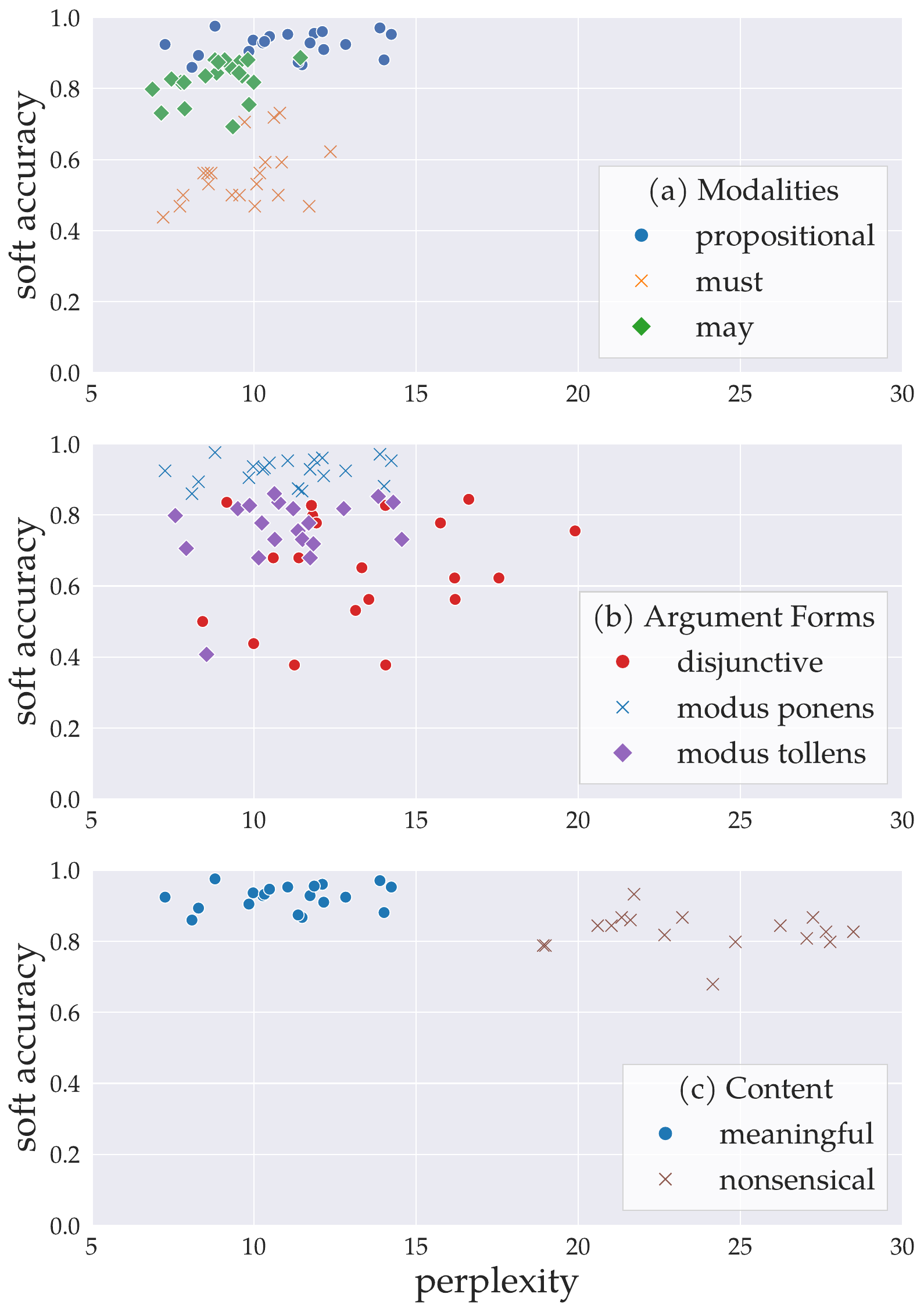}
    \vspace{-5pt}
    \caption{
        \label{fig:teaser}
        Illustration of the fact that perplexity does not serve as a reliable indicator of logical reasoning performance; and therefore, neither does probability.
        The distributions of the probabilities assigned to the ground-truth answer (i.e., soft accuracy; Y-axis) by Llama-3-70B are plotted against the perplexity of the corresponding example question (X-axis) and grouped by (a) modality, (b) argument forms, and (c) logic interpretation content.
        Each group consists of 20 randomly selected examples with other factors controlled.
    }
    \vspace{-5pt}
\end{figure}

Logical reasoning is a fundamental aspect of building AI systems for reliable decision-making \interalia{kautz-etal-1992-planning}---given a set of premises, an AI system should be able to deduce valid conclusions.
With the advent of large language models \citep[LLMs; ][\textit{inter alia}]{touvronLlamaOpenFoundation2023,jiangMistral7B2023a,metaLlamaHerdModels2024}, there has been a surge of interest in using these models to assist planning and decision-making \interalia{huang-etal-2022-language};
therefore, understanding the logical reasoning capabilities becomes crucial in understanding the reliability and potential of LLMs in planning.
While recent work has shown that LLMs exhibit decent performance on logical reasoning problems \interalia{liuLogiQAChallengeDataset2020a,ontanonLogicInferenceNewDatasaet2022,wanLogicAskerEvaluatingImproving2024},
there is still a lack of fine-grained understanding of the logical forms---among many argument forms presented in natural language \citep{shieber-1993-problem}, do LLMs perform equally well, or do they exhibit preferences for certain argument forms?
Do more complex components of logical forms, such as modalities, matter for LLM performance?

In this work, we investigate the logical reasoning capabilities of LLMs by assessing their performance on different logical forms.
We curate a dataset of natural language statements and questions based on several logical forms in both propositional and modal logic, which is designed to mirror reasoning in daily communication.
An example is shown in \cref{subsec:involved-logical-forms}.
We then conduct a series of controlled experiments to analyze the performance of a set of LLMs on the dataset.
Although our findings generally align with those by \citet{gonen-etal-2023-demystifying} and \citet{mccoyEmbersAutoregressionShow2024}, who suggest that LLMs excel on examples with high probability, our results indicate that logical form, including but not limited to modalities and argument forms, is a crucial complementary factor in predicting the performance of LLMs (\cref{fig:teaser}).
Additionally, with meaningful real-world interpretations, we find that:
\begin{enumerate}[topsep=2pt, itemsep=-3pt, parsep=0.5em, leftmargin=*]
    \item LLMs are still far from being perfect in atomic-level propositional and modal logic reasoning.
    \item LLMs prefer an affirmative answer under the modality of possibility, whereas they prefer a negative answer under the modality of necessity.
    \item In line with the recent results on categorical syllogisms \citep{eisape-etal-2024-systematic}, we verify on hypothetical and disjunctive syllogisms that LLMs achieve better performance on certain logical forms that humans perform well.
          However, some logical forms receive favor from LLMs, while the phenomena lack support from human intuition or human behavioral data.
\end{enumerate}
This paper is structured as follows.
After reviewing related work (\cref{sec:related}), we describe the dataset synthesis process (\cref{sec:dataset}).
We report the LLM reasoning results on our data (\cref{sec:experiment}) and compare them with human performance (\cref{sec:human}).
We conclude by discussing the implications of our results and the limitations(\cref{sec:discussion}).

%% file: chapters/02-related.tex
\section{Related Work}
\label{sec:related}
\noindent \textbf{Logical reasoning benchmarks.}
Existing LLM logical reasoning benchmarks \interalia{liuLogiQAChallengeDataset2020a,hanFolioNaturalLanguage2024} focus on complex, multi-hop reasoning problems with manually annotated problems, making cross-problem comparisons challenging.
Recent work has introduced benchmarks with synthesized natural-language questions using predefined logical formulas and substitution rules \interalia{saparovLanguageModelsAre2022, saparovTestingGeneralDeductive2023, parmar-etal-2024-logicbench,wanLogicAskerEvaluatingImproving2024}.
Compared to them, our work uniquely incorporates modal logic, which has been largely unexplored in existing benchmarks---while \citet{hollidayConditionalModalReasoning2024} present a case study, our approach offers two key advances: controlled knowledge bias in logic interpretations (\cref{subsec:involved-logical-forms}) and a more rigorous statistical evaluation framework (\cref{sec:experiment-measure}).

\vspace{2pt}
\noindent \textbf{Propositional and modal logic reasoning in language models.}
Recent work has explored training and finetuning language models specifically for logical reasoning \citep{clark-etal-2021-transformers,hahn-etal-2021-teaching,tafjord-etal-2022-entailer}.
Our work differs in two key aspects: (1) we evaluate general-purpose language models through prompting, a cost-efficient setup that has been widely adopted in recent years, and we focus on propositional and alethic modal logic rather than temporal \citep{hahn-etal-2021-teaching} or epistemic \citep{sileo-lernould-2023-mindgames} logic; \footnote{Technically, any logic that involves non-truth-functional operators, including first-order logic, temporal logic, and epistemic logic, can be viewed as a modal logic; however, we adopt the most restrictive sense of \textit{modal logic} \citep{sep-logic-modal-origins} and use it interchangeably with \textit{alethic modal logic}.}
(2) unlike studies comparing LLM and human performance on categorical syllogisms \interalia{eisape-etal-2024-systematic},\footnote{
    We refer readers to \citet{zong-lin-2024-categorical} for a more comprehensive review of categorical syllogisms.} we focus on hypothetical and disjunctive syllogisms with considerations of modality.

\vspace{2pt}
\noindent\textbf{Human logic reasoning.}
Work on human reasoning capabilities has informed studies of LLM logical reasoning: \citet{eisape-etal-2024-systematic} compared LLM syllogistic reasoning with human behavior results \citep{ragniWhenDoesReasoner2019} under the framework of the Mental Models Theory \citep{johnson-1983-mental}; \citet{lampinenLanguageModelsHumans2024} found similar content effects in human and LLM reasoning, supporting the need to control for common-sense knowledge in benchmarks (\cref{subsec:syn-natural-language});
\citet{belemPerceptionsLinguisticUncertainty2024a} studied human and LLM perception of uncertainty at a lexical level.
Compared to them, we focus on the propositional and modal logic reasoning process and contribute new behavioral data.

%% file: chapters/03-dataset.tex
\section{Dataset}
\label{sec:dataset}
\begin{figure*}[t]
  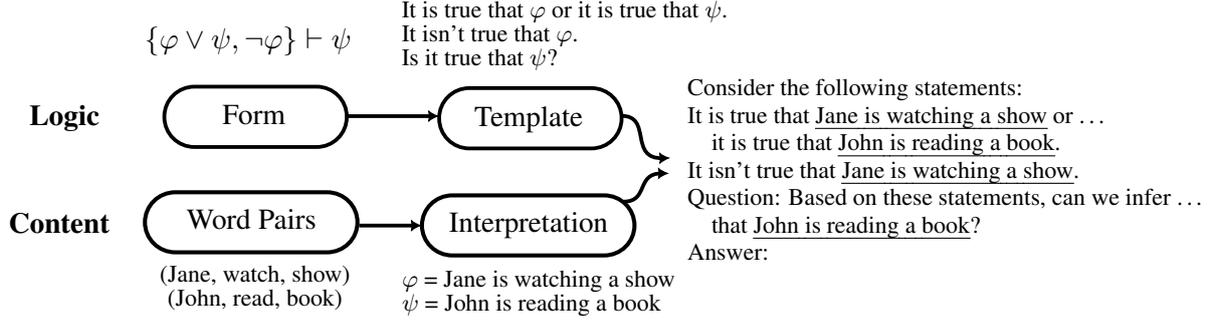
  \caption{
    The data synthesis pipeline: for each variable in logic forms (\cref{subsec:syn-logic}), we assign meanings to them to obtain the natural language question-answering pairs (\cref{subsec:syn-natural-language}).
  }
  \label{fig:syn-pipeline}
\end{figure*}

We curate a dataset of natural-language multi-choice questions to measure the logical inference performance of LLMs.
Starting from propositional and modal logical forms as templates (\cref{subsec:syn-logic}), we assign meanings (e.g., real-world interpretations) to each variable and translate templates into natural-language Yes/No questions (\cref{subsec:syn-natural-language}).
A subsidiary visualization of the process is shown in \cref{fig:syn-pipeline}.

\subsection{Background: Propositional and Modal Logic}
\label{subsec:syn-logic}

Propositional logic studies the relation between propositions.
In this framework, each proposition is typically represented by a variable, and multiple propositions combine with logical connectives (e.g., $\lor$ and $\rightarrow$) to form compound propositions.

In propositional logic, a proposition can be evaluated as either true or false; however, this system can be overly simplistic when dealing with the complexity of real-world events.
Consider the statement \textit{Alice is not eating}, while it is true in a world where Alice is not eating, it may become false in a hypothetical \textit{possible world} where Alice is indeed eating.
This idea, known as possible world semantics \citep{kripkeCompletenessTheoremModal1959}, provides a framework for more nuanced statements about event possibilities, such as \textit{Alice may be eating} and \textit{Alice must be eating}.
The former statement can be understood as there \textit{exists} a possible world where Alice is eating, and the latter can be understood as in \textit{all} possible worlds, Alice is eating.\footnote{
    The possible world semantics, therefore, connects the notion of \textit{necessity} and \textit{possibility} to the universal and existential quantification ($\forall, \exists$) under first-order logic.
}
Normal modal logic \citep{kripkeSemanticalAnalysisModal1963} formalizes this idea and extends propositional logic to reason about event necessity and possibility.
In the Backus--Naur form, a normal modal logic system $\mathcal{L}$ can be written as
\begin{align}
    \mathcal{L} : \varphi
    \coloneqq\; &
    p \mid
    \lnot \varphi  \mid
    \Box \varphi  \mid
    \Diamond \varphi  \mid \nonumber          \\
                & \varphi  \lor \varphi  \mid
    \varphi \land \varphi  \mid
    \varphi \to \varphi,
    \label{eqn:modal-logic-system-definition}
\end{align}
where $p$ is a propositional variable that serves as an atom in $\mathcal{L}$, $\lnot$ is the negation operator, $\Box$ is the necessity operator (\textit{must}), $\Diamond$ is the possibility operator (\textit{may}), $\lor$ is logical disjunction (\textit{or}), $\land$ is logical conjunction (\textit{and}), and $\to$ is the logical implication operator (\textit{if...then}).
$\varphi$ denotes the syntactic category of a formula in $\mathcal{L}$.
The right-hand side of \cref{eqn:modal-logic-system-definition} describes all possible logical formulas under the system $\mathcal{L}$: for example, if $\varphi \in \mathcal{L}$, the rules imply that $\lnot \varphi \in \mathcal{L}$, $\Box \varphi \in \mathcal{L}$, and so on.
Following the convention in logic, the operator precedence is $\{\lnot, \Box, \Diamond\} \succ \{\lor, \land\} \succ \{\to\}$.

Indeed, the operators $\left(\lnot, \Box, \to\right)$ forms a functional complete set of operators under $\mathcal{L}$.
Suppose $\varphi$ and $\psi$ are variables that represent logical formulas.
The logical or ($\lor$) and logical and ($\land$) operators can be rewritten with logical not ($\lnot$) and logical implication ($\to$), as follows:
\noindent
\begin{align}
    \varphi \lor \psi  & \Leftrightarrow \lnot \varphi \to \psi, \label{eqn:def-lor}          \\
    \varphi \land \psi & \Leftrightarrow \lnot \left(\varphi \to \lnot \psi\right). \nonumber
\end{align}
Possibility operator $\Diamond$ can also be derived from the necessity operator. \begin{align}\Diamond \varphi &\Leftrightarrow \lnot \Box \lnot \varphi\end{align}

\vspace{3pt}
\noindent\textbf{Deduction and sequent.}
Given a formula set $\Gamma$ as premises, if a deduction to a conclusion $\varphi$ exists using axiom schemata and inference rules under the normal modal logic, we say the premises \textit{infer} the conclusion, and the deduction can be represented as a logic \textit{sequent} \(\Gamma \vdash \varphi\).
If a formula set $\Gamma$ do not infer the conclusion, we denote it as \(\Gamma \nvdash \varphi\) and call it a \textit{non-entailment}.

\subsection{Translating Logic to Natural Language}
\label{subsec:syn-natural-language}
An \textit{interpretation} maps propositional variables to concrete meanings.
For example, under the interpretation that $p$ is ``\textit{Jane is eating apples}'' and $q$ is ``\textit{John is eating oranges}'', the logical formula $p \lor q$ becomes ``\textit{Jane is eating apples or John is eating oranges}.''

Choices of interpretation, i.e., the concrete content of the sentence, should not affect the underlying logical reasoning process.
However, in natural-language utterances, reasoning can be influenced by various confounding factors.
Knowledge bias is a common pitfall.
For example, given the logical form \(\left\{\Box p \to \Box \lnot q, \Box p\right\} \vdash \Box \lnot q\), regardless of $p$'s interpretation, if we interpret $\lnot q\coloneqq\textit{``Cats are not animals''}$ then the conclusion will be ``\textit{It is certain that cats are not animals.}''
But common-sense knowledge suggests that ``\textit{it is certain that cats are animals}'' ($\Box q$), which logically contradicts the existing premise set.\footnote{
    This confounding factor affects the examples in Table 9 of \citet{hanFolioNaturalLanguage2024}.
}
Such bias will complicate logical reasoning \citep{lampinenLanguageModelsHumans2024} and should be avoided in data curation.
Besides, each variable should have independent interpretation, as detailed in \cref{subsec:logic-translate-strategy}.

After being assigned interpretations, each logical form is further articulated as a yes-no question on whether the conclusion can be inferred from the premises.
To mitigate the ambiguity in natural language, we design heuristic rules to translate logic forms into less ambiguous English, which are detailed in \cref{subsec:logic-translate-strategy}.
For the exact wordings we used, see \cref{tab:question-full} in Appendices.
If a valid deduction exists (\(\vdash\)) for the logical form,
the ground truth answer is \texttt{Yes}, otherwise \texttt{No}.
The answer is solely determined by the logical form and is independent of the interpretation.

\subsection{Involved Logical Forms}
\label{subsec:involved-logical-forms}

Translated logical forms can have varying degrees of naturalness.
For example, the \textit{necessitation rule} \(\left\{\varphi\right\} \vdash \Box \varphi \), which translates to ``\(\varphi\) is true; therefore, it is certain that \(\varphi\) is true,'' appears to be unnatural due to redundancy.
\footnote{Nevertheless, we report the experiment results on necessitation rule in \cref{subsec:extra-intro-modality}.}
Based on the relationship between $\lor$ and $\to$ in \cref{eqn:def-lor}, we use hypothetical and disjunctive syllogisms with four basic variants:
\begin{align}
    \left\{\varphi \lor \psi, \lnot \varphi\right\} & \vdash \psi, \tag{$\lor^\mathrm{L}$} \label{eqn:inf-rule-or-left}                      \\
    \{\lnot \varphi \to \psi, \lnot \varphi\}       & \vdash \psi, \tag{$\to^{\mathrm{L}}$; modus ponens}   \label{eqn:inf-rule-imply-left}  \\
    \{\varphi \lor \psi , \lnot \psi\}              & \vdash \varphi \tag{$\lor^\mathrm{R}$}, \label{eqn:inf-rule-or-right}                  \\
    \{\lnot \varphi \to \psi, \lnot \psi\}          & \vdash \varphi \tag{$\to^\mathrm{R}$; modus tollens}. \label{eqn:inf-rule-imply-right}
\end{align}
Despite the semantic similarity, these logical forms translate to different natural-language questions.
For example, taking the interpretations of $\varphi := \textit{Jane is watching a show}$ and $\psi := \textit{John is reading a book}$, $\lor^\mathrm{L}$ translates to
\begin{table}[H]
    \vspace{-5pt}
    \begin{tabular}{p{0.92\linewidth}}
        Consider the following statements:                          \\
        \textit{Jane is watching a show or John is reading a book.} \\
        \textit{Jane isn't watching a show.}                        \\
        Question: Based on these statements, can we infer that \textit{John is reading a book}?
    \end{tabular}
    \vspace{-10pt}
\end{table}
\noindent With the same interpretation, $\to^\mathrm{L}$'s translation of the first statement is \textit{If Jane isn't watching a show, then John is reading a book.}

According to the commutativity of disjunction operator, we group $\lor^\mathrm{L}$ and $\lor^\mathrm{R}$ together as \textit{disjunctive syllogism}, alongside two hypothetical syllogism groups, modus ponens ($\to^\mathrm{L}$) and modus tollens ($\to^\mathrm{R}$).
All the logical forms shown above are valid sequents with ground-truth answer \texttt{Yes}.
To balance the dataset, we introduce some logic fallacies that generate questions with ground-truth label \texttt{No}.
By flipping the second premises and the conclusions, we obtain the following fallacies:
\begin{align*}
    \left\{\varphi \lor \psi, \psi\right\} & \nvdash \lnot \varphi \tag{$\lor^\mathrm{L}_{\nvdash}$}, \label{eqn:fallacy-inf-rule-or-left}       \\
    \{\lnot \varphi \to \psi, \psi\}       & \nvdash \lnot \varphi, \tag{$\to^{\mathrm{L}}_{\nvdash}$}   \label{eqn:fallacy-inf-rule-imply-left} \\
    \{\varphi \lor \psi , \varphi\}        & \nvdash \lnot \psi \tag{$\lor^\mathrm{R}_{\nvdash}$}, \label{eqn:fallacy-inf-rule-or-right}         \\
    \{\lnot \varphi \to \psi, \varphi\}    & \nvdash \lnot \psi \tag{$\to^\mathrm{R}_{\nvdash}$}, \label{eqn:fallacy-inf-rule-imply-right}
\end{align*}
where $\lor^\mathrm{L}_{\nvdash}$ and $\lor^\mathrm{R}_{\nvdash}$ are grouped as \textit{affirming the disjunction}, $\to^{\mathrm{L}}_{\nvdash}$ and $\to^\mathrm{R}_{\nvdash}$ corresponds to \textit{affirming the consequent} and \textit{denying the antecedent}, respectively.
In our dataset, we require the formulas \(\varphi\) and \(\psi\) to the form of \(\modal p\) and \(\modal q\),  where \(p\) and \(q\) are propositional variables, each assigned with an interpretation.
Both variables are constrained under the same modality \(\modal\), which can be necessity ($\Box$), possibility ($\Diamond$) or no modality ($\varnothing$).
Pairing with four rules and theorem--fallacy variations, we have a total of $3\times 4\times 2=24$ forms.

\subsection{Involved Logic Interpretations}

For logic interpretations, we generate a set of verb phrases by prompting the CodeLlama 2 model \citep{roziereCodeLlamaOpen2024}, and select 204 of them manually.
and combine them with top-200 popular baby names in the US into subject-verb-object pairs,\footnote{\url{https://www.ssa.gov/oact/babynames/names.zip}} such as \(\left(\textit{Ray}, \textit{make}, \textit{a pizza}\right)\).
We randomly generate $1000$ interpretations with two pairs each.
The same set of interpretations is applied to variables $p, q$ in each logic sequent's natural langauge template.
In total, there are $24\times 1000 = 24000$ question, with samples shown in \cref{tab:question-full}.

%% file: asset/figure/data-pipeline.pdf_tex
\begingroup%
  \makeatletter%
  \providecommand\color[2][]{%
    \errmessage{(Inkscape) Color is used for the text in Inkscape, but the package 'color.sty' is not loaded}%
    \renewcommand\color[2][]{}%
  }%
  \providecommand\transparent[1]{%
    \errmessage{(Inkscape) Transparency is used (non-zero) for the text in Inkscape, but the package 'transparent.sty' is not loaded}%
    \renewcommand\transparent[1]{}%
  }%
  \providecommand\rotatebox[2]{#2}%
  \newcommand*\fsize{\dimexpr\f@size pt\relax}%
  \newcommand*\lineheight[1]{\fontsize{\fsize}{#1\fsize}\selectfont}%
  \ifx\svgwidth\undefined%
    \setlength{\unitlength}{595.2756338bp}%
    \ifx\svgscale\undefined%
      \relax%
    \else%
      \setlength{\unitlength}{\unitlength * \real{\svgscale}}%
    \fi%
  \else%
    \setlength{\unitlength}{\svgwidth}%
  \fi%
  \global\let\svgwidth\undefined%
  \global\let\svgscale\undefined%
  \makeatother%
  \begin{picture}(1,0.22857141)%
    \lineheight{1}%
    \setlength\tabcolsep{0pt}%
    \put(0.15661387,0.13909194){\color[rgb]{0,0,0}\makebox(0,0)[t]{\lineheight{1.25}\smash{\begin{tabular}[t]{c}Form\end{tabular}}}}%
    \put(0,0){\includegraphics[width=\unitlength,page=1]{data-pipeline.pdf}}%
    \put(0.3288361,0.1381886){\color[rgb]{0,0,0}\makebox(0,0)[t]{\lineheight{1.25}\smash{\begin{tabular}[t]{c}Template\end{tabular}}}}%
    \put(0,0){\includegraphics[width=\unitlength,page=2]{data-pipeline.pdf}}%
    \put(0.32869437,0.0710278){\color[rgb]{0,0,0}\makebox(0,0)[t]{\lineheight{1.25}\smash{\begin{tabular}[t]{c}Interpretation\end{tabular}}}}%
    \put(0,0){\includegraphics[width=\unitlength,page=3]{data-pipeline.pdf}}%
    \put(0.1551089,0.07299545){\color[rgb]{0,0,0}\makebox(0,0)[t]{\lineheight{1.25}\smash{\begin{tabular}[t]{c}Word Pairs\end{tabular}}}}%
    \put(0,0){\includegraphics[width=\unitlength,page=4]{data-pipeline.pdf}}%
    \put(0.08823707,0.18704774){\color[rgb]{0,0,0}\makebox(0,0)[lt]{\lineheight{1}\smash{\begin{tabular}[t]{l}$\left\{\varphi \lor \psi, \lnot \varphi\right\} \vdash \psi$\end{tabular}}}}%
    \put(0.15731568,0.03960935){\small \color[rgb]{0,0,0}\makebox(0,0)[t]{\lineheight{1}\smash{\begin{tabular}[t]{c}(Jane, watch, show)\\(John, read, book)\end{tabular}}}}%
    \put(0.24936225,0.0363846){\small \color[rgb]{0,0,0}\makebox(0,0)[lt]{\lineheight{1}\smash{\begin{tabular}[t]{l}$\varphi$ = Jane is watching a show\\$\psi$ = John is reading a book\end{tabular}}}}%
    \put(0,0){\includegraphics[width=\unitlength,page=5]{data-pipeline.pdf}}%
    \put(0.24914377,0.20635735){\small \color[rgb]{0,0,0}\makebox(0,0)[lt]{\lineheight{1}\smash{\begin{tabular}[t]{l}It is true that $\varphi$ or it is true that $\psi$.\\It isn't true that $\varphi$.\\Is it true that $\psi$?\end{tabular}}}}%
    \put(0.42823769,0.15739721){\small \color[rgb]{0,0,0}\makebox(0,0)[lt]{\lineheight{1.15}\smash{\begin{tabular}[t]{l}Consider the following statements:\\It is true that \uline{Jane is watching a show} or \dots\\\quad it is true that \uline{John is reading a book}.\\ It isn't true that \uline{Jane is watching a show}.\\ Question: Based on these statements, can we infer \dots\\\quad that \uline{John is reading a book}?\\ Answer:\end{tabular}}}}%
    \put(0.01608858,0.13872497){\color[rgb]{0,0,0}\makebox(0,0)[lt]{\lineheight{1.25}\smash{\begin{tabular}[t]{l}\textbf{Logic}\end{tabular}}}}%
    \put(0.00302021,0.07044885){\color[rgb]{0,0,0}\makebox(0,0)[lt]{\lineheight{1.25}\smash{\begin{tabular}[t]{l}\textbf{Content}\end{tabular}}}}%
  \end{picture}%
\endgroup%

%% file: chapters/04-experiments.tex
\section{Experiment}
\label{sec:experiment}

\subsection{Metrics and Investigated Models}
\label{sec:experiment-measure}

\begin{table*}[!t]
    \centering
    \include{asset/table/tab-softacc-base}

    \caption{
        Overall and break-down accuracies of different models, as well as their HuggingFace OpenLLM Leaderboard performance and relative ranking \citep{open-llm-leaderboard-v2}.
        Each argument form category denotes the union of the fine-grained categories specified in the superscripts and subscripts---for example, $\lor^{\mathrm{L}, \mathrm{R}}_{\vdash}$ denotes the entire disjunctive syllogism group.
        \textbf{Boldfaced} values indicate the row-wise maximum for each factor.
        Note that due to technical limitations of commercial LLMs, results from OpenAI-o1 \citep{openai-o1} and Gemini-1.5-pro \citep{team2024gemini} are greedy-decoding based evaluation on 2,000 random samples that serve as references, and are therefore not directly comparable to other probability-based evaluations.
        Human results are detailed in \cref{sec:human}.
    }
    \label{tab:softacc-base}
\end{table*}

\citet{huPromptingNotSubstitute2023} have suggested that the standard approach of greedily decoding yes-no strings \citep{dentella-etal-2023-systematic} may underestimate the competence of a language model; therefore, we adopt a probability-based metric to evaluate the model performance.
In our evaluation protocol, the predicted likelihood of the tokens \texttt{Yes} and \texttt{No}, conditioned on the prompt \(s\)---denoted as \(p(\texttt{Yes}\mid s)\) and \(p(\texttt{No}\mid s)\), respectively---serve as the soft labels for yes-no answers.
The soft accuracy $\hat p$ on the single example with ground-truth answer $y\in \{\texttt{Yes}, \texttt{No}\}$ is defined as the relative probability of $y$:
\noindent
$$
    \hat p = \frac{p(\texttt{No} \mid s)\mathbbm{1}[y = \texttt{No}] + p(\texttt{Yes} \mid s) \mathbbm{1}[y = \texttt{Yes}]}{p(\texttt{No} \mid s) + p(\texttt{Yes} \mid s)},
$$
\noindent
where $\mathbbm{1}[\cdot]$ is the indicator function that returns 1 if the condition is true and 0 otherwise.
This relative probability can also be viewed as the confidence score of the model on the ground-truth answer.
The soft accuracy \(\mathit{Acc}_\mathrm{soft}\) of a model on the entire dataset $\mathcal{D}$ is defined as the average soft accuracy over all examples,
\begin{align*}
    \mathit{Acc}_\textit{soft} = \frac{1}{\lvert\mathcal{D}\rvert} \sum_{i=1}^{\lvert \mathcal{D}\rvert} \hat p_i.
\end{align*}
\vspace{-10pt}

\noindent We use a zero-shot setting to investigate the general performance of the models' logical inference capabilities---while adding detailed instructions or few-shot demonstrations may increase the absolute performance, they are at the cost of introducing possibly undesired confounding factors or behaviors, such as simply copy-pasting the answers in the examples.

We evaluate on the following models with open-sourced weights: mistral-7b-v0.2 and -8x7b \citep{jiangMistral7B2023a, jiangMixtralExperts2024};
llama-2-7b, -13b and -70b \citep{touvronLlamaOpenFoundation2023};
3.1 version of llama-3-8b and -70b \citep{metaLlamaHerdModels2024};
yi-34b \citep{yiOpenFoundation2024};
phi-2 and phi-3-mini \citep{javaheripiPhi2023, abdinPhi3TechnicalReport2024}.\footnote{
    Our evaluation protocol technically requires the conditional probabilities of specified answers given a prompt, which are not supported by most commercial models; however, we report the greedy-decoding accuracy of these models on a sample subset for reference.
}

\subsection{Results: Performance w.r.t. Logical Forms}
We evaluate the aforementioned models with the probability-based protocol (\cref{tab:softacc-base}).
Generally, models that rank higher in the leaderboard also achieve higher soft accuracy on our dataset.
The break-down accuracies on modalities and argument forms reveal that:

\begin{enumerate}[leftmargin=*,itemsep=0pt,topsep=2pt]
    \item (Modality) All models consistently perform better on the possibility ($\Diamond$) than necessity ($\Box$) or plain propositional logic.
    \item (Argument Forms) The pattern is more diverse, yet most of the models struggle the most on modus tollens ($\to^\mathrm{R}_\vdash$) within logic sequents (i.e., questions with ground-truth answers \texttt{Yes}), and affirming the consequent ($\to^\mathrm{L}_\nvdash$) within fallacies.
\end{enumerate}

\subsubsection{Analysis on Logic Sequents}
To systematically analyze the effect on model performance of each factor of interest, as well as cross-validating the observations above, we fit a linear mixed-effects model \citep{raudenbush-2002-hierarchical} to the soft accuracy data on valid logic sequents (i.e. with ground truth of \texttt{Yes}) across different LLMs and logical forms,
\begin{align}
    \mathit{Acc}_\textit{soft} & \sim \textit{Modality} + \textit{ArgForm} + \textit{Perplexity} \nonumber \\
                               & + (1 + \textit{Perplexity} \mid \textit{LLM}),
    \label{eqn: mixed-effects}
\end{align}
with the linear fixed effects of (i.) modality, (ii.) argument form, and (iii.) input perplexity.
Individual probability, coupled with a constant term, is modeled as a random effect to account for potential model-specific biases.
Here, \textit{Perplexity} denotes the perplexity of the input text ($x_1x_2\dots x_N$), which is defined as the exponential of the token-wise average negative log-likelihood of the text given a specific language model:
\begin{align*}
    \textit{Perplexity} = \exp\left(-\frac{1}{N}\sum_{i=1}^{N}\log p(x_i \mid x_{<i})\right)
\end{align*}
The mixed-effects model yields a marginal $R^2$ of $0.342$ and a conditional $R^2$ of $0.543$, suggesting a reasonable predictive power.
The likelihood ratio test on the full regression model vs. the null regression model without each of the fixed effects yields a significant result ($p<0.001$), suggesting the importance of all these factors in determining the model performance.

\begin{figure}[t]
    \centering
    \vspace{-5pt}
    \includegraphics[
        width=0.98\columnwidth,
        keepaspectratio,
    ]{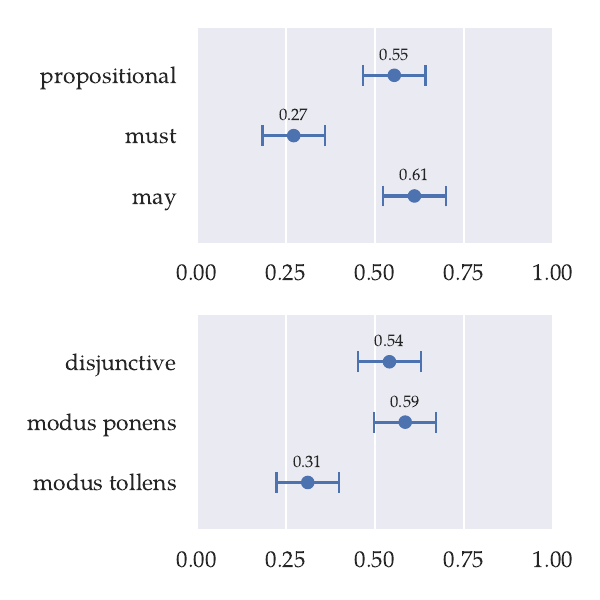}
    \vspace{-10pt}
    \caption{
        Estimated marginal means of logical form factors in the mixed-effects model of \cref{eqn: mixed-effects}, along with their 95\% confidence intervals.
    }
    \vspace{-5pt}
    \label{fig:emmeans-lm}
\end{figure}
\begin{table}[t]
    \centering \small
    \begin{tabular}{lc}
        \toprule
        \textbf{Hypothesis}                           & \textbf{$p$-value} \\
        \midrule
        $\text{propositional} < \text{may}$           & $<0.001$           \\
        $\text{must} < \text{propositional}$          & $<0.001$           \\
        $\text{must} < \text{may}$                    & $<0.001$           \\
        \midrule
        $\text{disjunctive} < \text{modus ponens}$    & $<0.001$           \\
        $\text{modus tollens} < \text{modus ponens}$  & $<0.001$           \\
        $\text{modus tollens} < \text{disjunctive}  $ & $<0.001$           \\
        \bottomrule
    \end{tabular}
    \caption{
        \label{tab:hypothesis-test}
        Hypothesis testing results on the effect of logical form factors on soft accuracy (\cref{fig:emmeans-lm}).
        \vspace{-5pt}
    }
\end{table}

\vspace{3pt}\noindent\textbf{Fixed effects.}
In line with \citet{gonen-etal-2023-demystifying} and \citet{mccoyEmbersAutoregressionShow2024}, we find a negative correlation between perplexity and soft accuracy ($p<0.001$); however, the correlation is weak ($\rho=-0.09$), which suggests the necessity of the complementary factors below in predicting LLM performance.

For different modalities and argument forms, we estimate their marginal means on soft accuracy (\cref{fig:emmeans-lm}) and perform pairwise hypothesis testing on the estimated coefficients (\cref{tab:hypothesis-test}).
The results generally align with the general observations on the full dataset.
The only exception is that modus ponens ($\to^\mathrm{L}$), instead of disjunctive syllogisms ($\lor$), appears to be the easiest argument form (i.e., the one with the highest soft accuracy) among all.

\begin{figure}[t]
    \centering
    \vspace{-10pt}
    \includegraphics[width=\linewidth]{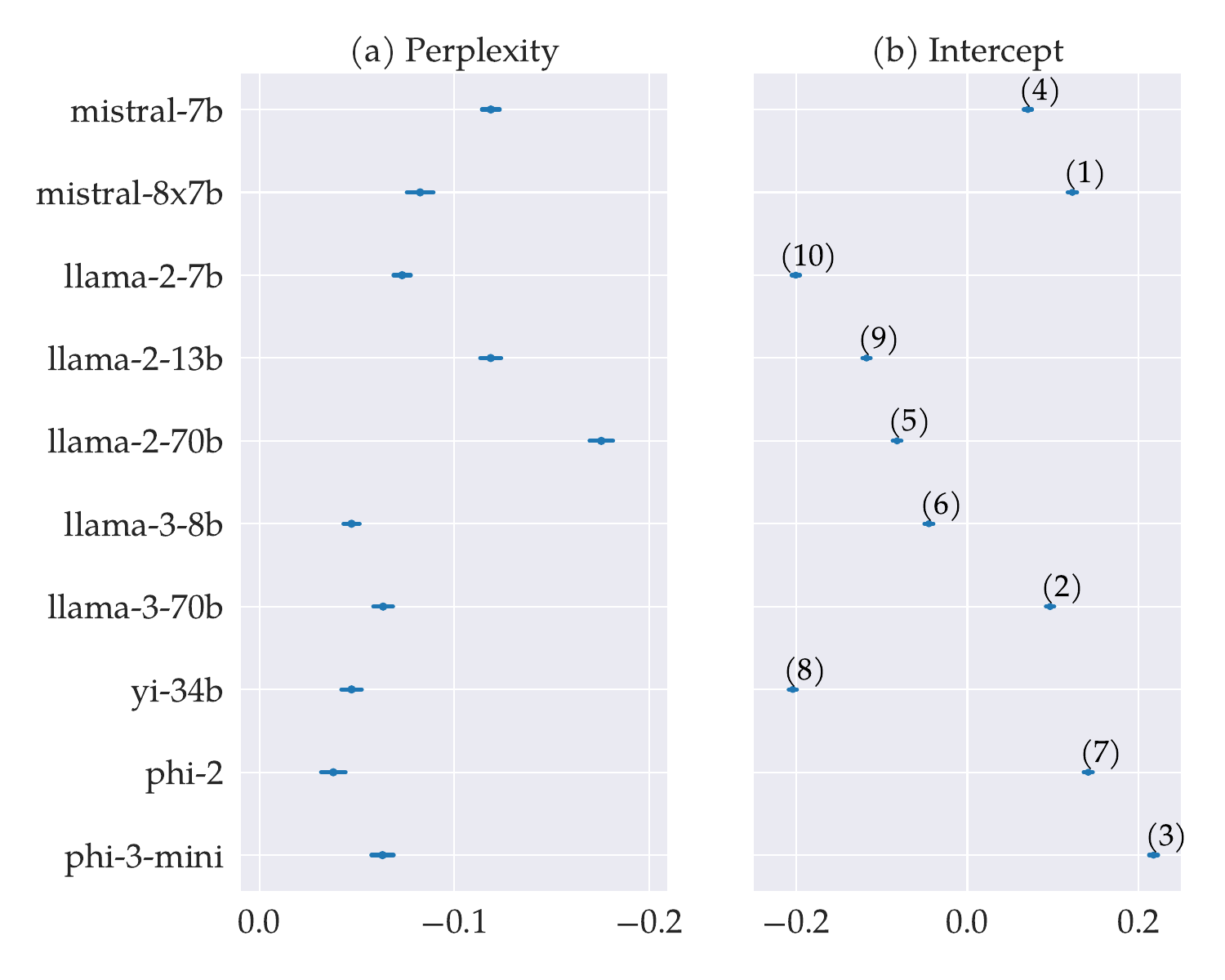}
    \vspace{-15pt}
    \caption{
        \label{fig:random-effects}
        Illustration of per-model random effects on soft accuracy in the mixed-effects model of \cref{eqn: mixed-effects} with 99.9\% confidence intervals.
        (a) Mixed effects (i.e., the sum of fixed and random effects) of perplexity. (b) Intercept random effects (i.e., constant term per model on soft accuracy), with the model performance rank (\cref{tab:softacc-base}) annotated in parentheses.
    }
    \vspace{-10pt}
\end{figure}
\begin{figure*}[t]
    \centering
    \vspace{-5pt}
    \includegraphics[width=1\textwidth,keepaspectratio]{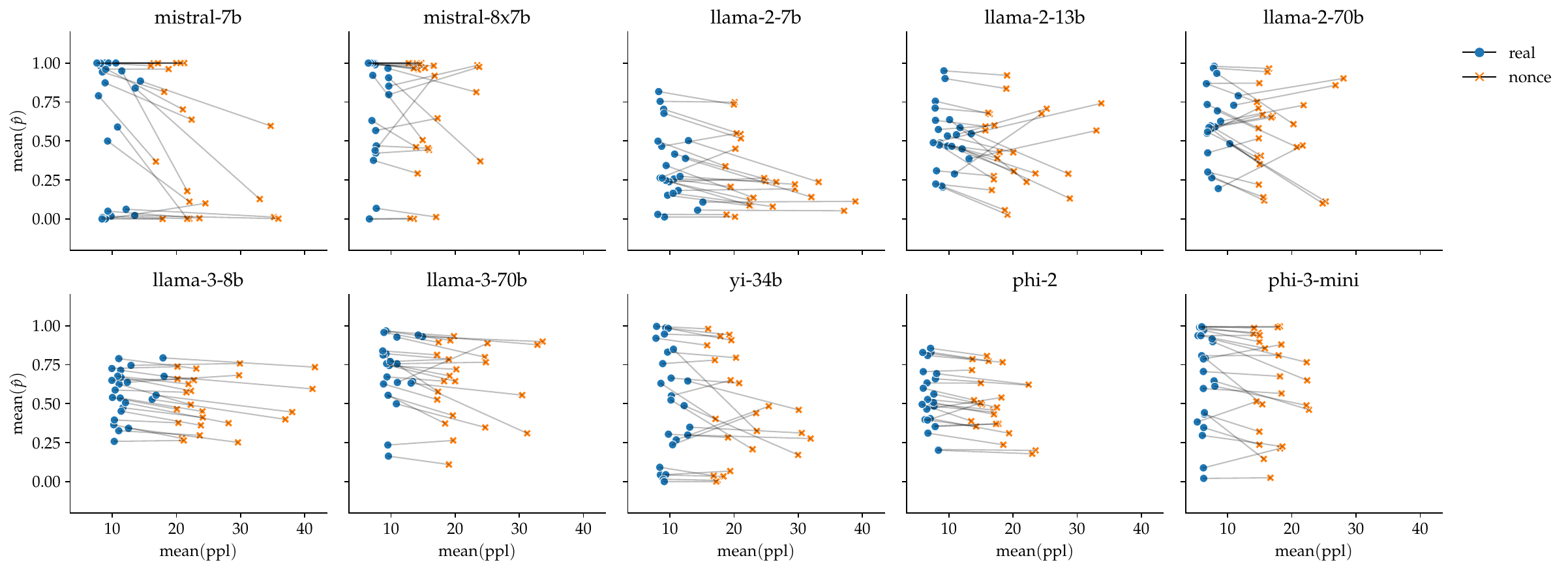}
    \vspace{-15pt}
    \caption{
        \label{fig:corr}
        Correlation between mean perplexity and mean confidence score on each logic sequent.
        Each point represents an average over a group of 1000 prompts that share the same underlying logic sequent.
        Two connected dots share the same logic formula.
    }
    \vspace{-10pt}
\end{figure*}

\vspace{3pt}\noindent\textbf{Random effects.}
We analyze the per-LLM random effects on the soft accuracy (\cref{fig:random-effects}).
All the model-specific mixed effects of perplexity are negative, suggesting the negative correlation between perplexity and soft accuracy is consistent across models (\cref{fig:random-effects}a).
While the intercept random effects are not perfectly aligned with the model performance---since the perplexity random effects may introduce confounding factors---higher-ranked models generally tend to have higher intercept random effects (\cref{fig:random-effects}b), which cross-validates the general performance ranking.

\subsubsection{Extended Analysis on the Negative Perplexity--Performance Correlation}
\label{sec:perplexity}
We further investigate the negative correlation between perplexity and model performance through a controlled experiment: we create a mirror dataset of the same size, keeping all the logical formulas while interpreting them with nonsensical words.
For example, the formula $\Diamond(\varphi \lor \psi)$ may be interpreted as \textit{it's possible that Neva is \uline{balaring} a \uline{montery} or Lucille is \uline{sweeling prandates}}, where the underlined words and phrases are nonsensical.
Intuitively, the perplexity of the problems in this mirror dataset should be much higher than that of the primary dataset problems (\cref{sec:dataset}) under any reasonably trained language model.

We analyze the correlation between perplexity and model performance (\cref{fig:corr}).
As desired, the perplexity of problems with nonsensical words are indeed much higher than that of the primary dataset ($\approx 30$ vs. $\approx 10$).
The significant portion of horizontal and inclined lines in the figures again suggests that perplexity is not a reliable predictor of model performance.
Meanwhile, the overall parallelism of the lines echos our results that logical forms are important factors for such prediction.

\begin{figure}[t]
    \centering
    \vspace{-5pt}
    \includegraphics[width=0.95\columnwidth,keepaspectratio]{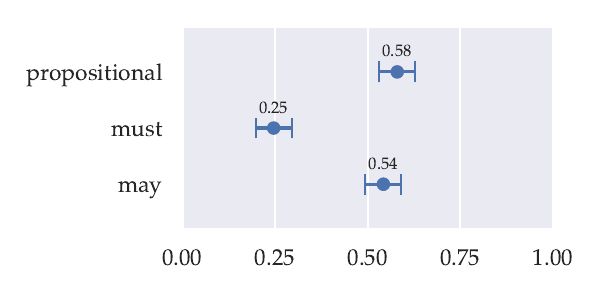}
    \vspace{-15pt}
    \caption{
        \label{fig:affirmation-rejection}
        Estimated marginal means of the factors in the mixed-effects model of \cref{eqn: mixed-effects-yesno} with 95\% confidence intervals. Higher coefficients indicate a higher tendency to affirm the claim.
    }
    \vspace{-10pt}
\end{figure}
\subsubsection{The Affirmation Bias over Modalities}
\label{subsec:affirmation-bias}

One key argument of \citet{dentella-etal-2023-systematic} is that large language models exhibit a bias towards affirming the claim, i.e., answering \texttt{Yes} more frequently than \texttt{No}.
We investigate this phenomenon by fitting a mixed-effects model
\begin{align}
     & \frac{P(\texttt{Yes}\mid s)}{P(\texttt{Yes} \mid s) + P(\texttt{No} \mid s)} \sim  \textit{Modality} + \textit{ArgForm} \nonumber \\
     & + \textit{Perplexity} + (1 + \textit{Perplexity} \mid \textit{LLM}),
    \label{eqn: mixed-effects-yesno}
\end{align}
which has the same structure as \cref{eqn: mixed-effects}, except the dependent variable being the relative probability of answering \texttt{Yes} conditioned on input text $s$.

We present the estimated marginal means of the factors in the mixed-effects model (\cref{fig:affirmation-rejection}).
While our results confirm the affirmation bias on propositional logic, such bias is slightly less pronounced on the possibility modality ($\Diamond$, around 0.03), and the models even show a bias towards rejecting claims under the necessity modality ($\Box$).

%% file: asset/table/tab-softacc-base.tex
\small
\setlength{\tabcolsep}{5pt}
\begin{tabular}{
    @{}lllllccccccccc@{}
    }
    \toprule
                   & \multicolumn{2}{c}{Overall} & \multicolumn{2}{c}{Leaderboard} & \multicolumn{3}{c}{Modality} & \multicolumn{6}{c}{Argument Form}                                                                                                                                                                                                                                                           \\
    \cmidrule(lr){6-8} \cmidrule(lr){9-14}
    Model          & \multicolumn{2}{c}{(Rank)}  & \multicolumn{2}{c}{(Rank)}      & $\varnothing$                & $\Box$                            & $\Diamond$     & $\lor^{\mathrm{L},\mathrm{R}}_{\vdash}$ & $\to^\mathrm{L}_{\vdash}$ & $\to^\mathrm{R}_{\vdash}$ & $\lor^{\mathrm{L},\mathrm{R}}_{\nvdash}$ & $\to^\mathrm{L}_{\nvdash}$ & $\to^\mathrm{R}_{\nvdash}$                                   \\
    \midrule
    mistral-7b     & 0.645                       & (4)                             & 0.145                        & (7)                               & 0.464          & 0.496                                   & \textbf{0.974}            & 0.877                     & 0.663                                    & 0.280                      & 0.434                      & 0.653          & \textbf{0.939} \\
    mistral-8x7b   & 0.724                       & (1)                             & 0.193                        & (5)                               & 0.698          & 0.601                                   & \textbf{0.874}            & \textbf{0.963}            & 0.873                                    & 0.023                      & 0.757                      & 0.648          & 0.813          \\
    llama-2-7b     & 0.335                       & (10)                            & 0.094                        & (10)                              & 0.262          & 0.207                                   & \textbf{0.538}            & 0.444                     & 0.147                                    & 0.315                      & 0.208                      & 0.451          & \textbf{0.468} \\
    llama-2-13b    & 0.513                       & (9)                             & 0.110                        & (9)                               & 0.488          & 0.362                                   & \textbf{0.688}            & 0.418                     & 0.581                                    & 0.393                      & \textbf{0.631}             & 0.436          & 0.591          \\
    llama-2-70b    & 0.611                       & (5)                             & 0.127                        & (8)                               & 0.616          & 0.471                                   & \textbf{0.746}            & 0.446                     & \textbf{0.845}                           & 0.518                      & 0.775                      & 0.389          & 0.694          \\
    llama-3-8b     & 0.565                       & (6)                             & 0.239                        & (3)                               & 0.598          & 0.460                                   & \textbf{0.639}            & 0.526                     & 0.470                                    & 0.332                      & 0.664                      & 0.625          & \textbf{0.716} \\
    llama-3-70b    & 0.714                       & (2)                             & 0.362                        & (1)                               & 0.745          & 0.554                                   & \textbf{0.843}            & 0.606                     & 0.773                                    & 0.515                      & \textbf{0.882}             & 0.661          & 0.788          \\
    yi-34b         & 0.518                       & (8)                             & 0.226                        & (4)                               & 0.457          & 0.413                                   & \textbf{0.683}            & 0.346                     & 0.498                                    & 0.205                      & 0.685                      & 0.638          & \textbf{0.737} \\
    phi-2          & 0.532                       & (7)                             & 0.155                        & (6)                               & 0.469          & 0.456                                   & \textbf{0.673}            & 0.670                     & \textbf{0.757}                           & 0.522                      & 0.365                      & 0.402          & 0.510          \\
    phi-3-mini     & 0.690                       & (3)                             & 0.272                        & (2)                               & 0.657          & 0.536                                   & \textbf{0.877}            & 0.839                     & \textbf{0.974}                           & 0.475                      & 0.664                      & 0.462          & 0.604          \\
    \cmidrule{1-14}
    OpenAI-o1      & 0.926                       & N/A                             & N/A                          & N/A                               & \textbf{1.000} & 0.773                                   & \textbf{1.000}            & 0.895                     & \textbf{1.000}                           & 0.775                      & 0.919                      & \textbf{1.000} & \textbf{1.000} \\
    Gemini-1.5-Pro & 0.859                       & N/A                             & N/A                          & N/A                               & 0.831          & 0.748                                   & 0.997                     & \textbf{1.000}            & \textbf{1.000}                           & 0.919                      & 0.661                      & 0.991          & 0.638          \\
    \cmidrule{1-14}
    human          & 0.595                       & N/A                             & N/A                          & N/A                               & 0.589          & 0.566                                   & \textbf{0.640}            & 0.691                     & \textbf{0.901}                           & 0.628                      & 0.594                      & 0.225          & 0.411          \\
    \bottomrule
\end{tabular}

%% file: chapters/05-human.tex
\section{Human Experiments}
\label{sec:human}

LLMs are trained on text produced by humans and are able to generate plausible text; therefore, there have been interests in using LLMs as human models \interalia{eisape-etal-2024-systematic,misra-kim-2024-generating}.
Following this line of work, we conduct a human behavioral experiment to ground the LLM reasoning behavior.
Using samples from our primary dataset, we collected 710 responses from adults fluent in English through Prolific.\footnote{\url{https://prolific.com}}
More experiment details can be found in \cref{subsec:human-details}.

The average human accuracy on each group is shown in the last row of \cref{tab:softacc-base}.\footnote{Human responses are binary classes, so correct and incorrect responses are coded as $1$ and $0$, respectively.}
Aligned with our LLM results (\cref{sec:experiment}), on modalities, the overall human results also show an accuracy order of ($\Diamond \succ \varnothing \succ \Box$),
and on argument forms, modus ponens ($\to^\mathrm{L}$) is the most accurately answered pattern.

\begin{figure}[!t]
    \centering
    \vspace{-5pt}
    \includegraphics[
        width=0.95\columnwidth,
        keepaspectratio,
    ]{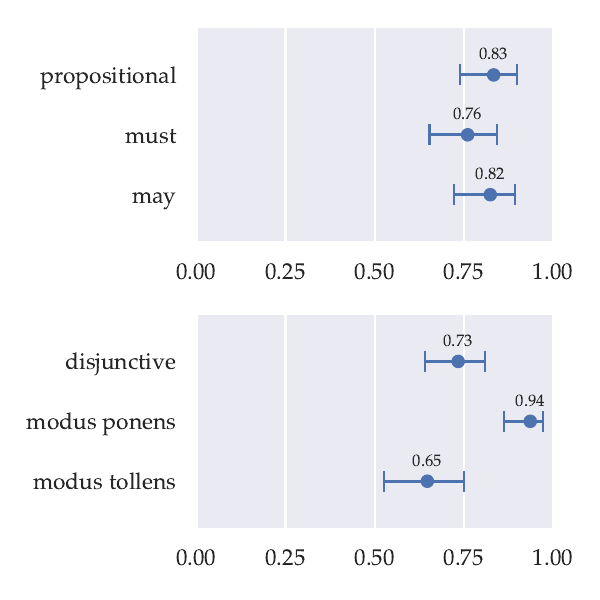}
    \vspace{-5pt}
    \caption{
        Estimated marginal means of logical form factors in the generalized mixed-effects model of \cref{eqn:mixed-effects-human}, along with their 95\% confidence intervals.
        \label{fig:emmeans-human}
    }
    \vspace{-10pt}
\end{figure}

To further investigate the interactions of logic factors, we fit a generalized linear mixed-effects model \citep{batesFittingLinearMixedEffects2015} to verify the effect of modality and argument forms on human logic reasoning accuracy (\cref{eqn:mixed-effects-human} and \cref{fig:emmeans-human}).
\noindent
\begin{align}
    \mathrm{logit}(\mathit{Acc}) & \sim \textit{Modality} + \textit{ArgForm} + \textit{Rt} \nonumber \\
                                 & + (1 + \textit{Rt} \mid \textit{ParticipantID}),
    \label{eqn:mixed-effects-human}
\end{align}
\noindent
where $\mathit{Acc}$ is the binary accuracy of human responses, and $\textit{Rt}$ is the response time.
The generalized mixed-effects model yields a marginal $R^2$ of $0.121$ yet a $0.419$ conditional $R^2$, indicating a diverse response pattern across participants.
The likelihood ratio test on the full model against the null model shows that only the effect of argument form is significant ($\chi^2(2)=25.6$, $p<0.001$).
However, in accordance with the overall performance, we find modus ponens ($\to^\mathrm{L}$) has a significantly higher effect than the other two valid argument forms.
This confirms that logical forms can also have a significant impact on human reasoning accuracy, which is consistent with the LLM results, although the effect sizes are not the same.

%% file: chapters/06-discussion.tex
\section{Conclusion and Discussion}
\label{sec:discussion}
We present an analysis of hypothetical and disjunctive syllogisms on propositional and modal logic and systematically analyze the LLM performance on the dataset.
Our analysis provides novel insights on explaining and predicting LLM performance: in addition to the perplexity or probability of the input text, the underlying logic forms play an important role in determining the performance of LLMs.
In addition, we compare the behaviors of LLMs and humans using the same data through human behavioral experiments.
We discuss the implications of our results as follows.

\vspace{2pt}
\noindent\textbf{Probability in language models.}
Probability and perplexity are often used as intrinsic evaluation metrics for language models.
While \citet{gonen-etal-2023-demystifying} and \citet{mccoyEmbersAutoregressionShow2024} show that probability and perplexity correlate well with LLM performance, literature in program synthesis with LLMs shows little correlation between probability and execution-based evaluation results \citep{li2022competition,shi-etal-2022-natural}.
This work does not necessarily contradict either line but rather provides complementary factors for analyzing LLM performance.

We argue that probability may have become an overloaded term in analyzing LLMs.
Low probability may be due to one or more of the following non-exhaustive reasons: (1) out-of-context content, (2) ungrammatical language, or (3) grammatical but semantically awkward content (cf. the mirror dataset in \cref{sec:perplexity}), (4) reasonable but rare content.
We hypothesize that the probability of language models may not be essentially able to capture all these nuanced differences, and call for encoding and decoding algorithms---such as \citet{meister-etal-2023-locally}---that can better decompose the probability into finer-grained and explainable components.

\vspace{2pt}
\noindent\textbf{Comparing humans and LLMs.}
What is our goal for building LLMs?
To achieve better performance on practical tasks or to build a more human-like model?
Our results, together with \citet{eisape-etal-2024-systematic}, suggest that these two goals may not be perfectly aligned by revealing a mixture of similarity and discrepancy between LLMs and humans---for example, while LLMs exhibit higher benchmark performance than humans on our dataset and show the same argument form preferences with humans (\cref{fig:emmeans-lm,fig:emmeans-human}), they also show systematic biases that we do not find significant in human reasoning (e.g., disfavoring the necessity modality, \cref{subsec:affirmation-bias}).
While there has been positive evidence of using LLMs as human models in psycholinguistic studies \interalia{misra-kim-2024-generating}, our results suggest executing such approaches cautiously.

\vspace{2pt}
\noindent\textbf{On the relation between modality and performance.}
Our results show that there is a significant difference in performance between necessity and possibility modalities, with the former much lower than the latter (\cref{tab:softacc-base}).
Part of the reason for this is that LLMs have a significant tendency to say ``No'' to the necessity modality (\cref{fig:affirmation-rejection}).

On the one hand, our results extend the conclusion of \citet{dentella-etal-2023-systematic} that LLMs generally respond positively---LLM behaviors may be significantly affected by finer-grained factors, including but not necessarily limited to the modality involved in the input.
On the other hand, while LLMs systematically tend to answer ``No'' to questions in necessity modality, we do not find related evidence in human experiments, which leads us to hypothesize that such rejection bias comes from either the model architecture or the training strategies, such as the reinforcement learning with human feedback \citep[RLHF;][]{ouyang-etal-2022-training} protocol.
We leave this as an open question for future research.

\vspace{2pt}
\noindent\textbf{Modal logic and theory of mind.}
Modality, in principle, encodes mental states and beliefs.
The reasoning of beliefs also resonates with the theory of mind \interalia{premackDoesChimpanzeeHave1978,baron-cohenDoesAutisticChild1985} and machine theory of mind \interalia{rabinowitzMachineTheoryMind2018, maHolisticLandscapeSituated2023}.
Following the effort by \citet{sileo-lernould-2023-mindgames} that uses epistemic modal logic to model the machine theory of mind, our work assesses the behaviors of LLMs on alethic modal logic, distantly revealing the future potential of LLMs in achieving the theory of mind.

%% file: chapters/07-closing.tex
\section*{Limitations}
This work comes with two major limitations:
\begin{enumerate}[leftmargin=*,topsep=0pt,itemsep=0pt]
      \item While we have verified that our data has a low perplexity ($9.82\pm 2.47$ under mistral-7b; much lower than that of the data by \citet{wanLogicAskerEvaluatingImproving2024}, $25.44$), and, therefore, are similar enough to natural language utterances, the synthetic language cannot fully substitute natural language in daily life.
            Our dataset and analysis are not comprehensive enough to cover many nuanced examples that may appear in real communication, especially when context-dependent understanding is crucial to conveying communication goals.
      \item Despite more than 7,000 languages worldwide, as a first step, our material only covers English.
            This narrow focus is due to the languages the authors are proficient in and the coverage of the language models.
            We acknowledge the importance of extending the scope of this work to a more comprehensive set of languages and leave the extension as an immediate follow-up step.
\end{enumerate}

In addition, the sample size of human experiments is somewhat limited.
We leave more comprehensive human behavioral data collection and analysis to future work.

\section*{Ethics Statement}

While this work involves human logical reasoning experiments, we have ensured that (1) the data are generated procedurally following templates listed in the paper and (2) there is no harmful content in the atomic logical interpretations, reviewed by all the authors.
In addition, we have ensured that all participants are paid a fair wage through the Prolific platform.
Instructions and consent forms delivered to the participants can be found in the \cref{subsec:human-details}.
The institutional ethics review board has approved the data collection process.

This work contributes to the understanding of LLMs.
We do not foresee risk beyond the minimal risk posed by LLM evaluation work.
We acknowledge that using LLMs in real-world scenarios could significantly impact human behaviors, raising the need for model transparency, safety, security, and interpretability.
We will open-source the synthetic logical reasoning dataset upon publication.

\section{Acknowledgements}
We thank Yudong Li for his help in setting up the Gemini and OpenAI API for the experiments.
This work was supported in part by a Google PhD Fellowship and a Canada CIFAR AI Chair award to FS, as well as NSERC RGPIN-2024-04395.

%% file: chapters/a1-responsible.tex
\section{Additional Experiment Details}

\subsection{LLM Experiment Details}
\label{subsec:llm-details}
All LLMs used are obtained from \href{https://huggingface.co/models}{Hugging Face} checkpoints.
Time and compute power requirements vary, the largest llama-3-70b model takes around 2 hours on NVIDIA A6000 GPU to obtain all results in \cref{sec:experiment}.

\subsection{Human Experiment Details}
\label{subsec:human-details}
\paragraph{Participant instructions.}
We use keys \textit{F} and \textit{J}, which are roughly symmetric on a standard English keyboard, to collect participant responses.
Half of the participants see the following instruction:

\textit{In this study, you will be presented with two statements followed by a question. Your task is to answer either Yes or No to the question, based on the information provided in the statements.
    Please respond quickly and accurately by pressing "F" for Yes, and "J" for No.}

To mitigate the possible bias introduced by the dominant hand, we have the other half of the participants see instruction with reversed keys:

\textit{In this study, you will be presented with two statements followed by a question. Your task is to answer either Yes or No to the question, based on the information provided in the statements.
    Please respond quickly and accurately by pressing "F" for No, and "J" for Yes.}

\paragraph{Participant wage.}
We offer participants an hourly wage of 1.5 times Prolific's minimum wage.
The duration is determined by the median completion time among all participants.

%% file: chapters/a2-sample.tex
\setcounter{table}{0}
\setcounter{figure}{0}
\setcounter{equation}{0}
\renewcommand{\thetable}{A\arabic{table}}
\renewcommand{\thefigure}{A\arabic{figure}}
\renewcommand{\theequation}{A\arabic{equation}}

\section{Extra Details of the Dataset}

\subsection{Considerations in Translating Logical Form to Natural Language}
\label{subsec:logic-translate-strategy}

During the interpretation process, another key point is to assign independent interpretations to variables.
Deciding the dependency also involves common sense knowledge.
For example, consider the premises $\lnot p \to q$ and $q$.
If we interpret $p \coloneqq\textit{``Jane is inside the house''}$ and $q\coloneqq\textit{``Jane is out''}$ to proposition variables $p$ and $q$, the two variables are possibly not independent.
According to common sense, ``\textit{Jane is not inside the house}'' ($\lnot p$) correlates with or is even equivalent to ``\textit{Jane is out}'' ($q$).
Logically, $\left\{\lnot p \to q, q\right\} \nvdash \lnot p$; however, with the extra premise $\lnot p \leftrightarrow q$ given by common sense, people may conclude that $\lnot p$.\footnote{
  This confounding factor affects the examples in Appendix C.1.12 of \citet{hollidayConditionalModalReasoning2024}.
}

Besides, natural language is ambiguous---one sentence in natural language can come from multiple logical forms under the same interpretation.
We use present tense and progressive aspect to encourage a reading of imaginary ongoing events, corresponding to the alethic modality.
Such events are less likely to induce LLM's or human's individual bias, as they are unrelated to factual knowledge or moral judgements.
Also, we always use two full verb phrases, ruling out sentences like ``\textit{Jane is eating apples or oranges,}'' so the two events are less likely to be mutually exclusive.
In this way, we can reduce the ambiguity of the questions in our dataset.

\subsection{Data Samples}

All logic forms and corresponding natural language sentences can be found in \cref{tab:question-full}.

The exact prompt format is as follows:

\begin{table}[H]
    \small
    \begin{tabular}{p{\linewidth}}
    Consider the following statements:\verb|\n|\\
    \uline{Jane is watching a show or John is reading a book.}\verb|\n|\\
    \uline{Jane isn't watching a show}.\verb|\n|\\
    Question: Based on these statements, can we infer that \uline{John is reading a book}?\verb|\n|\\
    Answer:\verb|<eof|>
    \end{tabular}
\end{table}
\vspace{-1em}

\newcommand{\subjectA}{\uline{Jane}}
\newcommand{\vpA}{\uline{watching a show}} 
\newcommand{\subjectB}{\uline{John}}
\newcommand{\vpB}{\uline{reading a book}}

\begin{table*}
\scriptsize
\include{asset/table/tab-question-full}
\caption{\textbf{Samples of all logical forms and corresponding natural language sentences.}}
\label{tab:question-full}
\end{table*}

%% file: asset/table/tab-question-full.tex
\begin{tabular}{
  @{}ccclp{0.52\textwidth}@{}
  }
\vspace{0.5em}
\textbf{Validity} & \textbf{Modality} & \textbf{Argument Form} & \textbf{Logical Form} & \textbf{Natural Language} \\
$\vdash$
& $\varnothing$ & $\lor^{\mathrm{L}}$ & $\{p \lor q, \lnot p\} \vdash q$ & 
   \subjectA{} is \vpA{} or \subjectB{} is \vpB{}.\newline
  \subjectA{} isn't \vpA{}.\newline
  Can we infer that \subjectB{} is \vpB{}? \\
& $\varnothing$ & $\lor^{\mathrm{R}}$ & $\{p \lor q, \lnot q\} \vdash p$ & 
   \subjectA{} is \vpA{} or \subjectB{} is \vpB{}.\newline
  \subjectB{} isn't \vpB{}.\newline
  Can we infer that \subjectA{} is \vpA{}? \\
& $\varnothing$ & $\rightarrow^{\mathrm{L}}$ & $\{\lnot p \to q, \lnot p\} \vdash q$ & 
   If \subjectA{} isn't \vpA{}, then \subjectB{} is \vpB{}.\newline
  \subjectA{} isn't \vpA{}.\newline
  Can we infer that \subjectB{} is \vpB{}? \\
& $\varnothing$ & $\rightarrow^{\mathrm{R}}$ & $\{\lnot p \to q, \lnot q\} \vdash p$ & 
   If \subjectA{} isn't \vpA{}, then \subjectB{} is \vpB{}.\newline
  \subjectB{} isn't \vpB{}.\newline
  Can we infer that \subjectA{} is \vpA{}? \\
& $\Box$ & $\lor^{\mathrm{L}}$ & $\{\Box p \lor \Box q, \lnot \Box p\} \vdash \Box q$ & 
   It's certain that \subjectA{} is \vpA{} or it's certain that \subjectB{} is \vpB{}.\newline
  It's uncertain whether \subjectA{} is \vpA{}.\newline
  Can we infer that it's certain that \subjectB{} is \vpB{}? \\
& $\Box$ & $\lor^{\mathrm{R}}$ & $\{\Box p \lor \Box q, \lnot \Box q\} \vdash \Box p$ & 
   It's certain that \subjectA{} is \vpA{} or it's certain that \subjectB{} is \vpB{}.\newline
  It's uncertain whether \subjectB{} is \vpB{}.\newline
  Can we infer that it's certain that \subjectA{} is \vpA{}? \\
& $\Box$ & $\rightarrow^{\mathrm{L}}$ & $\{\lnot \Box p \to \Box q, \lnot \Box p\} \vdash \Box q$ & 
   If it's uncertain whether \subjectA{} is \vpA{}, then it's certain that \subjectB{} is \vpB{}.\newline
  It's uncertain whether \subjectA{} is \vpA{}.\newline
  Can we infer that it's certain that \subjectB{} is \vpB{}? \\
& $\Box$ & $\rightarrow^{\mathrm{R}}$ & $\{\lnot \Box p \to \Box q, \lnot \Box q\} \vdash \Box p$ & 
   If it's uncertain whether \subjectA{} is \vpA{}, then it's certain that \subjectB{} is \vpB{}.\newline
  It's uncertain whether \subjectB{} is \vpB{}.\newline
  Can we infer that it's certain that \subjectA{} is \vpA{}? \\
& $\Diamond$ & $\lor^{\mathrm{L}}$ & $\{\Diamond p \lor \Diamond q, \lnot \Diamond p\} \vdash \Diamond q$ & 
   It's possible that \subjectA{} is \vpA{} or it's possible that \subjectB{} is \vpB{}.\newline
  It's impossible that \subjectA{} is \vpA{}.\newline
  Can we infer that it's possible that \subjectB{} is \vpB{}? \\
& $\Diamond$ & $\lor^{\mathrm{R}}$ & $\{\Diamond p \lor \Diamond q, \lnot \Diamond q\} \vdash \Diamond p$ & 
   It's possible that \subjectA{} is \vpA{} or it's possible that \subjectB{} is \vpB{}.\newline
  It's impossible that \subjectB{} is \vpB{}.\newline
  Can we infer that it's possible that \subjectA{} is \vpA{}? \\
& $\Diamond$ & $\rightarrow^{\mathrm{L}}$ & $\{\lnot \Diamond p \to \Diamond q, \lnot \Diamond p\} \vdash \Diamond q$ & 
   If it's impossible that \subjectA{} is \vpA{}, then it's possible that \subjectB{} is \vpB{}.\newline
  It's impossible that \subjectA{} is \vpA{}.\newline
  Can we infer that it's possible that \subjectB{} is \vpB{}? \\
& $\Diamond$ & $\rightarrow^{\mathrm{R}}$ & $\{\lnot \Diamond p \to \Diamond q, \lnot \Diamond q\} \vdash \Diamond p$ & 
   If it's impossible that \subjectA{} is \vpA{}, then it's possible that \subjectB{} is \vpB{}.\newline
  It's impossible that \subjectB{} is \vpB{}.\newline
  Can we infer that it's possible that \subjectA{} is \vpA{}? \\

$\nvdash$
& $\varnothing$ & $\lor^{\mathrm{L}}$ & $\{p \lor q, q\} \nvdash \lnot p$ & 
   \subjectA{} is \vpA{} or \subjectB{} is \vpB{}.\newline
  \subjectB{} is \vpB{}.\newline
  Can we infer that \subjectA{} isn't \vpA{}? \\
& $\varnothing$ & $\lor^{\mathrm{R}}$ & $\{p \lor q, p\} \nvdash \lnot q$ & 
   \subjectA{} is \vpA{} or \subjectB{} is \vpB{}.\newline
  \subjectA{} is \vpA{}.\newline
  Can we infer that \subjectB{} isn't \vpB{}? \\
& $\varnothing$ & $\rightarrow^{\mathrm{L}}$ & $\{\lnot p \to q, q\} \nvdash \lnot p$ & 
   If \subjectA{} isn't \vpA{}, then \subjectB{} is \vpB{}.\newline
  \subjectB{} is \vpB{}.\newline
  Can we infer that \subjectA{} isn't \vpA{}? \\
& $\varnothing$ & $\rightarrow^{\mathrm{R}}$ & $\{\lnot p \to q, p\} \nvdash \lnot q$ & 
   If \subjectA{} isn't \vpA{}, then \subjectB{} is \vpB{}.\newline
  \subjectA{} is \vpA{}.\newline
  Can we infer that \subjectB{} isn't \vpB{}? \\
& $\Box$ & $\lor^{\mathrm{L}}$ & $\{\Box p \lor \Box q, \Box q\} \nvdash \lnot \Box p$ & 
   It's certain that \subjectA{} is \vpA{} or it's certain that \subjectB{} is \vpB{}.\newline
  It's certain that \subjectB{} is \vpB{}.\newline
  Can we infer that it's uncertain whether \subjectA{} is \vpA{}? \\
& $\Box$ & $\lor^{\mathrm{R}}$ & $\{\Box p \lor \Box q, \Box p\} \nvdash \lnot \Box q$ & 
   It's certain that \subjectA{} is \vpA{} or it's certain that \subjectB{} is \vpB{}.\newline
  It's certain that \subjectA{} is \vpA{}.\newline
  Can we infer that it's uncertain whether \subjectB{} is \vpB{}? \\
& $\Box$ & $\rightarrow^{\mathrm{L}}$ & $\{\lnot \Box p \to \Box q, \Box q\} \nvdash \lnot \Box p$ & 
   If it's uncertain whether \subjectA{} is \vpA{}, then it's certain that \subjectB{} is \vpB{}.\newline
  It's certain that \subjectB{} is \vpB{}.\newline
  Can we infer that it's uncertain whether \subjectA{} is \vpA{}? \\
& $\Box$ & $\rightarrow^{\mathrm{R}}$ & $\{\lnot \Box p \to \Box q, \Box p\} \nvdash \lnot \Box q$ & 
   If it's uncertain whether \subjectA{} is \vpA{}, then it's certain that \subjectB{} is \vpB{}.\newline
  It's certain that \subjectA{} is \vpA{}.\newline
  Can we infer that it's uncertain whether \subjectB{} is \vpB{}? \\
& $\Diamond$ & $\lor^{\mathrm{L}}$ & $\{\Diamond p \lor \Diamond q, \Diamond q\} \nvdash \lnot \Diamond p$ & 
   It's possible that \subjectA{} is \vpA{} or it's possible that \subjectB{} is \vpB{}.\newline
  It's possible that \subjectB{} is \vpB{}.\newline
  Can we infer that it's impossible that \subjectA{} is \vpA{}? \\
& $\Diamond$ & $\lor^{\mathrm{R}}$ & $\{\Diamond p \lor \Diamond q, \Diamond p\} \nvdash \lnot \Diamond q$ & 
   It's possible that \subjectA{} is \vpA{} or it's possible that \subjectB{} is \vpB{}.\newline
  It's possible that \subjectA{} is \vpA{}.\newline
  Can we infer that it's impossible that \subjectB{} is \vpB{}? \\
& $\Diamond$ & $\rightarrow^{\mathrm{L}}$ & $\{\lnot \Diamond p \to \Diamond q, \Diamond q\} \nvdash \lnot \Diamond p$ & 
   If it's impossible that \subjectA{} is \vpA{}, then it's possible that \subjectB{} is \vpB{}.\newline
  It's possible that \subjectB{} is \vpB{}.\newline
  Can we infer that it's impossible that \subjectA{} is \vpA{}? \\
& $\Diamond$ & $\rightarrow^{\mathrm{R}}$ & $\{\lnot \Diamond p \to \Diamond q, \Diamond p\} \nvdash \lnot \Diamond q$ & 
   If it's impossible that \subjectA{} is \vpA{}, then it's possible that \subjectB{} is \vpB{}.\newline
  It's possible that \subjectA{} is \vpA{}.\newline
  Can we infer that it's impossible that \subjectB{} is \vpB{}? \\
\end{tabular}

%% file: chapters/a3-extra.tex
\section{Additional Experiments}

\subsection{Extra Experiment: Introduction Rule of Modality}
\label{subsec:extra-intro-modality}

We report the results on the necessitation rule and its variants here, as these rules are obscure and verbose to be articulated in natural language:
\begin{align}
    \left\{\varphi\right\} &\vdash \Box \varphi, \tag{necessitation rule} \label{eqn:necessitation-rule} \\
    \left\{\varphi\right\} &\vdash \Diamond \varphi, \nonumber \\
    \left\{\varphi\right\} &\vdash \varphi. \nonumber
\end{align}
Its natural language form is as follows:
\begin{table}[H]
    \small
    \begin{tabular}{rp{0.8\linewidth}}
    & Jane is watching a show.\\
    ($\Box$) & Can we infer that it's certain that Jane is watching a show?\\
    ($\Diamond$) & Can we infer that it's possible that Jane is watching a show?\\
    ($\varnothing$) & Can we infer that Jane is watching a show?\\
    \end{tabular}
\end{table}

All three variants are paired with 1000 logic interpretations.
As they are all rules of inference, the ground truth answer is always \texttt{Yes}.
Overall accuracy is shown in \cref{tab:softacc-necessitate},
where across all LLMs, the necessitation rule has the lowest accuracy.
This echoes the necessity modality's tendency to be rejected discussed in \cref{subsec:affirmation-bias}.

\begin{table}[t]
    \include{asset/table/tab-softacc-necessitate}
    \caption{
        \label{tab:softacc-necessitate}
        Overall accuracy of the necessitation rule and its modality variants on each model.
    }
\end{table}
    
We further fit a linear mixed-effects model similar to \cref{eqn: mixed-effects}, except that the argument form effect is now constant across all data points.
The mixed-effects model yields a marginal $R^2$ of $0.391$ and a conditional $R^2$ of $0.745$.
Estimated marginal means shows that the accuracy on $\varnothing$ is $0.171$ less than $\Diamond$, but $0.371$ higher than $\Box$, with both differences significant at $p < 0.0001$.
This further suggests that modality serves as an important factor on logic reasoning performance.

\subsection{Extra Experiment: Distribution of Modalities}

Besides the necessitation rule, \textit{distribution axiom} is the other fundamental axiom in normal modal logic.
It can be transformed into the rule shown in \cref{eqn:distribution-must-axiom},
and plugging in the definition of $\lor$ in \cref{eqn:def-lor} gives the rule shown in \cref{eqn:distribution-or-theorem}.
Notice that \cref{eqn:distribution-or-theorem} closely resembles rule \ref{eqn:inf-rule-or-left}'s variant with necessity, as shown in \cref{eqn:inf-rule-or-left-must},
except the different scope of the necessity operator and the position of the negation operator.
Moving the negation operator out of the necessity operator will result in a fallacy (Eq. \ref{eqn:distribution-or-fallacy}).
\noindent
\begin{align}
    \left\{\Box(\varphi \to \psi), \Box \varphi\right\} &\vdash \Box \psi, 
    \label{eqn:distribution-must-axiom} \\
    \left\{\Box(\varphi \lor \psi), \Box \lnot \varphi\right\} &\vdash \Box \psi \label{eqn:distribution-or-theorem}, \\
    \left\{\Box \varphi \lor \Box \psi, \lnot \Box \varphi\right\} &\vdash \Box \psi \label{eqn:inf-rule-or-left-must}, \\
    \left\{\Box(\varphi \lor \psi), \lnot \Box \varphi\right\} &\nvdash \Box \psi \label{eqn:distribution-or-fallacy}.
\end{align}
\noindent
We say \cref{eqn:distribution-or-theorem,eqn:inf-rule-or-left-must,eqn:distribution-or-fallacy} are of argument form \texttt{theorem}, \texttt{base} and \texttt{spurious}, respectively.
See \cref{tab:extra-distribution-forms} for the logical forms and their ground truth we used to study the distribution of modalities.
The natural language form is as follows:
\begin{table}[H]
    \small
    \begin{tabular}{rp{0.65\linewidth}}
    & It's certain that if Freddy is not going shopping, then Coy is making dinner.\\
    (\texttt{theorem}) & It's certain that Freddy is not going shopping.\\
    (\texttt{spurious}) & It's uncertain whether Freddy is going shopping.\\
    & Can we infer that it's certain that Coy is making dinner?\\
    \end{tabular}
\end{table}

This group of rules and fallacies comes from the fact that the necessity modality $\Box$ is not distributive to disjunction, i.e. $\Box (\varphi \lor \psi) \nvdash \Box \varphi \lor \Box \psi$ \citep[Ex. 5]{xiang-2019a-two-types}.
In contrast, the possibility modality $\Diamond$ is distributive to disjunction.
This particular case could have served as a material to test the LLM's knowledge of the asymmetry between the two modalities,
yet in \cref{subsec:affirmation-bias} we showed that there is a bias towards rejection on the necessity modality.
As the false case of the disjunction is on the necessity modality, this bias confounds the experiment.

\begin{table}[t]
    \include{asset/table/tab-distribution-axiom}
    \caption{
        \label{tab:extra-distribution-forms}
        Logical forms and their ground truth to study the distribution of modalities.
        Only the spurious form of the necessity modality (marked by \uline{underline}) has a ground truth of false.
    }
\end{table}

We fit a linear mixed-effects model similar to \cref{eqn: mixed-effects} to the data,
\noindent
\begin{align}
    \mathit{Acc}_\textit{soft} & \sim \textit{Modality} \times \textit{ArgForm} + \textit{Perplexity} \nonumber \\
                               & + (1 + \textit{Perplexity} \mid \textit{LLM}), \nonumber
\end{align}
\noindent
with an interaction term between the modality and argument form.
On the \texttt{theorem} form compared to the \texttt{base} form, the necessity modality $\Box$ has a $0.173$ higher estimated marginal means with $p < 0.0001$ significance, yet the possibility modality $\Diamond$ has a $0.071$ lower estimated marginal means.
On the \texttt{spurious} form compared to the \texttt{base} form, the $\Box$ has a $0.312$ higher means, and the $\Diamond$ has no significant difference.
On both forms, $\Diamond \succ \Box$ in terms of accuracy still holds at a slight margin of $0.110$ and $0.047$ respectively.

To verify whether on $\Box$ the performance increase on \texttt{spurious} form is due to the rejection bias, we fit a linear mixed-effects model with the relative probability of answering \texttt{Yes} as dependent variable.
Results show that on \texttt{spurious} form compared to the \texttt{base} form, the effect of $\Box$'s tendency to answer \texttt{Yes} is only $0.060$ lower, indicating the rejection bias of the \texttt{base} form is still present.
Therefore, we hypothesize that the LLM's performance on recognizing the fallacy of necessity distribution over disjunction is hindered by the rejection bias on the necessity modality.

%% file: asset/table/tab-softacc-necessitate.tex
\small
\centering
\begin{tabular*}{0.85\columnwidth}{
 @{\extracolsep{\fill}}lccc@{}
 }
 \toprule
  & $\varnothing$ & $\Box$ & $\Diamond$ \\
 \midrule
    mistral-7b & 0.998 & 0.885 & 0.999\\
    mistral-8x7b & 0.957 & 0.540 & 0.987\\
    llama-2-7b & 0.768 & 0.013 & 0.920\\
    llama-2-13b & 0.368 & 0.004 & 0.829\\
    llama-2-70b & 0.511 & 0.051 & 0.834\\
    llama-3-8b & 0.398 & 0.225 & 0.783\\
    llama-3-70b & 0.674 & 0.384 & 0.794\\
    yi-34b & 0.960 & 0.382 & 0.999\\
    phi-2 & 0.814 & 0.226 & 0.892\\
    phi-3-mini & 0.992 & 0.925 & 0.994\\
 \bottomrule
\end{tabular*}

%% file: asset/table/tab-distribution-axiom.tex
\small\centering
\begin{tabular}{ccl}
    \toprule
Modality & Argument Form & Logical Form \\
    \midrule
$\varnothing$ & base & $\varphi \lor \psi, \lnot \varphi \vdash \psi$ \\
$\Box$ & base & $\Box \varphi \lor \Box \psi, \lnot \Box \varphi \vdash \Box \psi$ \\
$\Box$ & theorem & $\Box ( \varphi \lor \psi), \Box \lnot \varphi \vdash \Box \psi$ \\
$\Box$ & \uline{spurious} & $\Box ( \varphi \lor \psi), \lnot \Box \varphi \nvdash \Box \psi$ \\
$\Diamond$ & base & $\Diamond \varphi \lor \Diamond \psi, \lnot \Diamond \varphi \vdash \Diamond \psi$ \\
$\Diamond$ & theorem & $\Diamond ( \varphi \lor \psi), \Diamond \lnot \varphi \vdash \Diamond \psi$ \\
$\Diamond$ & spurious & $\Diamond ( \varphi \lor \psi), \lnot \Diamond \varphi \vdash \Diamond \psi$ \\
    \bottomrule
    \end{tabular}